\documentclass{llncs}  

\usepackage{graphicx} 
\usepackage{subfig} 
\usepackage[T1]{fontenc}
\usepackage{amsmath}
\usepackage[english]{babel}
\usepackage{array,tabularx}
\usepackage{booktabs} 
\usepackage{makeidx}  
\usepackage{xparse}
\usepackage{moredefs}
\usepackage{lips}
\usepackage{epstopdf} 
\usepackage{color}
\usepackage{csquotes}
\usepackage[table,xcdraw]{xcolor}
\usepackage{times}
\usepackage[hidelinks]{hyperref}
\usepackage[nolist,nohyperlinks]{acronym}
\usepackage{comment}
\usepackage{paralist}
\usepackage{cleveref}
\usepackage{wrapfig}
\usepackage{multirow}
\usepackage{microtype}
\usepackage{listings}
\usepackage[super]{nth}
\usepackage{todonotes}
\usepackage{lscape}
\usepackage[normalem]{ulem}
\useunder{\uline}{\ul}{}
\usepackage{multirow}

\usepackage{pifont}
\newcommand{\xmark}{\ding{55}}%

\usepackage{float}
\floatstyle{plaintop}
\restylefloat{table}

\usepackage[style=base,labelfont=bf,labelsep=period,tableposition=top,font=small]{caption}
\makeatletter
\renewcommand\@biblabel[1]{#1.}
\makeatother
\addto\captionsenglish{}

\usepackage{fancyhdr}
\fancypagestyle{WI_footer}{\fancyhf{}\fancyfoot[L]{\color{black}\scriptsize 17th International Conference on Wirtschaftsinformatik\\February 2022, Nürnberg, Germany}}

\usepackage{cite}

\crefformat{footnote}{#2\footnotemark[#1]#3}
\expandafter\def\expandafter\UrlBreaks\expandafter{\UrlBreaks
  \do\a\do\b\do\c\do\d\do\e\do\f\do\g\do\h\do\i\do\j%
  \do\k\do\l\do\m\do\n\do\o\do\p\do\q\do\r\do\s\do\t%
  \do\u\do\v\do\w\do\x\do\y\do\z\do\A\do\B\do\C\do\D%
  \do\E\do\F\do\G\do\H\do\I\do\J\do\K\do\L\do\M\do\N%
  \do\O\do\P\do\Q\do\R\do\S\do\T\do\U\do\V\do\W\do\X%
  \do\Y\do\Z}

\newcolumntype{L}[1]{>{\raggedright\arraybackslash}p{#1}}   
\newcolumntype{C}[1]{>{\centering\arraybackslash}p{#1}}     
\newcolumntype{R}[1]{>{\raggedleft\arraybackslash}p{#1}}    

\begin{document}
\frontmatter          

\mainmatter              

\title{A Light in the Dark: Deep Learning Practices for Industrial Computer Vision}



\author{Maximilian Harl\inst{1}$^*$ \and
Marvin Herchenbach\inst{1}$^*$ \and
Sven Kruschel\inst{1}$^*$ \and
Nico Hambauer\inst{1}$^*$ \and
Patrick Zschech\inst{1} \and
Mathias Kraus\inst{1}}

\institute{Friedrich-Alexander University Erlangen-Nuremberg, Institute for Information Systems, Nuremberg, Germany\\
\email{\{maximilian.harl, marvin.herchenbach, nico.hambauer, sven.kruschel, patrick.zschech, mathias.kraus\}@fau.de}
\def\thefootnote{*}\footnotetext{These authors contributed equally to this work}\def\thefootnote{\arabic{footnote}}
}







\maketitle
\setcounter{footnote}{0}

\begin{abstract}
In recent years, large pre-trained deep neural networks (DNNs) have revolutionized the field of computer vision (CV). Although these DNNs have been shown to be very well suited for general image recognition tasks, application in industry is often precluded for three reasons: 1) large pre-trained DNNs are built on hundreds of millions of parameters, making deployment on many devices impossible, 2) the underlying dataset for pre-training consists of general objects, while industrial cases often consist of very specific objects, such as structures on solar wafers, 3) potentially biased pre-trained DNNs raise legal issues for companies. As a remedy, we study neural networks for CV that we train from scratch. For this purpose, we use a real-world case from a solar wafer manufacturer. We find that our neural networks achieve similar performances as pre-trained DNNs, even though they consist of far fewer parameters and do not rely on third-party datasets. 

\end{abstract}

\thispagestyle{WI_footer}


\section{Introduction}
\label{sec:introduction}

Deep learning (DL) has recently received increasing attention in research and industry alike, where we can find manifold examples of successful applications like machine translation \cite{singh_machine_2017}, predictive maintenance \cite{rezaeianjouybari_deep_2020}, autonomous driving \cite{friederich2020review}, and predictive business process monitoring \cite{harl2020explainable}. Given a sufficiently large amount of training data, DL has the advantage of automatically detecting useful patterns from raw input data without the need of manually defining sophisticated features for prediction purposes \cite{kraus_deep_2020, janiesch_machine_2021}.

A field that significantly benefits from DL functionality is that of computer vision (CV). It seeks to automatically extract useful information from images to mimic human capabilities of visual perception \cite{sager_survey_2021, szeliski_computer_2010}. On this basis, time-consuming and labor-intensive tasks like the recognition, detection, localization, tracking, and counting of objects can be supported more efficiently to save resources and relieve the burden of human workers \cite{zschech_picture_2021}.

Fueled by the broad access to computing power and the availability of large image repositories, the field of CV is currently experiencing a considerable phase of scientific progress and dissemination. We can observe that a global community of developers provides user-friendly programming frameworks, shares reusable software code, and builds huge pre-trained DL models based on large-scale image repositories \cite{chollet_deep_2018, liu_deep_2020, zschech_picture_2021}. This is beneficial for many applications, since the weights of pre-trained DL models can be reused via transfer learning, which helps to build models with limited training data, save computational resources, and avoid architecture search \cite{kurkova_survey_2018}.

However, relying on pre-trained models is not always a valid option. Especially in highly regulated settings, such as manufacturing environments, pre-trained models based on public image data pose a serious security risk due to e.g. adversarial attacks \cite{heinrich_fool_2020} or induced biases \cite{janiesch_machine_2021}. Similarly, public images usually contain generic objects such as houses, cars, and animals, which rarely reflect the shapes of objects in industrial applications like technical components and surface defects \cite{zschech_mit_2020}. Beyond that, pre-trained models often consist of several million parameters \cite{heinrich_is_2019}, which limits their application on small devices with limited hardware. In an era of edge computing (i.e., computations close to the source of data, usually in real-time), these limitations become even more decisive as large models naturally lead to longer inference times. 

As a remedy, this paper studies DL-based CV models for image recognition that are created from scratch for specific business needs. Based on a real-world case from a manufacturer of solar panels, we demonstrate how our lightweight models provide competitive prediction results for the task of automated quality inspection compared to large pre-trained models. To carry out our research, we conduct a series of computational experiments and comparative evaluation studies to provide guidance for practitioners and researchers on how to successfully develop DL models from scratch.

Our paper is structured as follows: In section \ref{sec:background}, we provide an overview about the conceptual background and related work, followed by a brief description of our industrial case in section \ref{sec:case}. We then proceed to elaborate on our experimental setup in section \ref{sec:experiments} by outlining our evaluation procedures and algorithmic approaches which can basically be structured into model-centric and data-centric development steps. Subsequently, we present our results in a comprehensive manner in section \ref{sec:results}. To conclude our paper, we provide summarizing remarks and derive an outlook for future work in section \ref{sec:conclusion}.

\section{Conceptual Background and Related Work}
\label{sec:background}

\subsection{Computer Vision and Deep Neural Networks}
\label{sec:CV-DNN}

The field of CV is concerned with the acquisition, processing, analysis, and understanding of digital images to generate symbolic or numerical information that can be used to support automated decision-making.
Just as humans use their eyes and brains to understand the world around them, CV attempts to produce the same effect so that computers can perceive and understand an image or a sequence of images and act accordingly in each situation.
This understanding can be achieved by disentangling high-level, symbolic information from low-level image features using models built with the help of geometry, statistics, physics, and learning theory \cite{sager_survey_2021}. Driven by academic and industrial motives, grand advances have been made in several areas such as 3D model building, optical character recognition, motion capture, disease diagnostics and surveillance \cite{szeliski_computer_2010}.

Nowadays, CV tasks are increasingly performed by machine learning (ML). Of particular interest are artificial neural networks (ANNs). Inspired by information processing in biological systems, they consist of multiple interconnected neural processing units that forward signals using weights and activation functions.
ANNs learn by processing many examples and iteratively adjusting the internal weights (= trainable parameters) according to the difference between predicted outcomes and the ground truth \cite{janiesch_machine_2021, goodfellow_deep_2016}.
In advanced approaches, neural processing units are typically organised into multiple layers that can be arranged into deep network architectures. 
This gives them the capability to process spatial information in raw image data and automatically extract features that are relevant for the prediction task, which is commonly known as deep learning \cite{lecun2015deep}.

\subsection{Convolutional Neural Networks and Transfer Learning}
\label{sec:CNN-TL}

A widely used network architecture in the field of CV is that of convolutional neural networks (CNNs) \cite{lecun2015deep}. They comprise a series of stages that allow hierarchical feature learning which is a useful principle to condense low-level features into higher-level concepts. For the task of object recognition, this means that the first few layers of the network are responsible for extracting basic features in the form of edges and corners. These are then incrementally aggregated into more complex features in the last few layers resembling the actual objects of interest, such as faces, cars, or defects.
Subsequently, these auto-generated features are used for prediction purposes to recognise objects of interest \cite{goodfellow_deep_2016}.

Over time, the CV community has developed increasingly deeper CNN architectures (e.g., VGG16 \cite{simonyan}) to benefit from the functionality of hierarchical feature learning. At the same time, more advanced mechanisms like inception modules (e.g., GoogleNet \cite{szegedy_going_2015}) and residual blocks (e.g., ResNet \cite{he2016deep}) have been introduced to improve the efficiency of deep multi-layered networks.
Most of these developments are based on large scale image repositories such as ImageNet \cite{deng_imagenet_2009} or COCO \cite{fleet_microsoft_2014} that contain generic objects like cars or houses. This yields the advantage that already trained models with their adjusted weights can be reused for further recognition tasks in similar domains, which is commonly known as transfer learning \cite{pan_survey_2010, kurkova_survey_2018}. For this purpose, pre-trained models can be utilized in two different ways: 1) either as generic feature extractor where most of the weights are frozen and only the last classification layers are adjusted for the task at hand, or 2) the weights of the feature extractor and classification layers are fine-tuned on new image-label-pairs using the initial weights as starting point in case that the source domain and the target domain differ too much from each other \cite{shaha_transfer_2018}.

The key advantage of transfer learning is that it greatly simplifies the development of CV models by requiring fewer computational resources, less training time, and much smaller amounts of training data \cite{kurkova_survey_2018}.
By contrast, building DNNs from scratch usually demands well-defined development steps that cover a mixture of \textit{model-centric} design choices (e.g., choice of architecture, setting of hyperparameters, choice of regularization, etc.) vs. \textit{data-centric} design choices (e.g., data augmentation, noise removal, improvement of label quality, etc.). For this reason, transfer learning is often the preferred choice in research and industry when developing DL-based systems for CV tasks \cite{rawat2017deep}.

However, for domains with specific business needs such as given in many industrial applications, pre-trained models also bear several risks. Firstly, large pre-trained DNNs are usually built on hundreds of millions of parameters \cite{heinrich_is_2019} that prevent an deployment on many edge computing devices due to limited storage capacities and the requirement of low inference times. Secondly, industrial cases usually consist of very specific objects, such as structures on surfaces or specific technical components that heavily deviate from the shapes of general training objects of public image repositories. Lastly, using pre-trained models from foreign sources can also pose a security risk as the models can be subject to biases and adversarial attacks \cite{janiesch_machine_2021, heinrich_fool_2020}. The latter have the potential to trick and break DNNs without obvious indication which could be demonstrated, for example, with perturbations of street signs in the field of autonomous driving \cite{eykholt_robust_2018}.

\section{Case Description}
\label{sec:case}

Our case refers to the automation of quality control in the photovoltaic (PV) industry. While automated quality control is common in many industries \cite{milo_new_2015, yang_using_2020, kurdyukova_development_2020}, the PV industry is lacking behind because of complex production environments. Consequently, the PV industry often still relies on manual human controls of PV wafers which is both tedious and costly and therefore offers grand potentials for improvements via automated inspections based on DL models \cite{deitsch_segmentation_2021, zschech_mit_2020}.

For this project, we partnered with a PV wafer manufacturer. 
To highlight damages in wafers, voltage is applied to make the wafers glow in the infrared spectral range and an infrared sensitive camera takes images. This leads to the effect that faulty areas appear darker on the wafer. 
Next, these images are labeled according to their error to obtain our dataset that is used to train a neural network that is able to classify wafer images.

\begin{figure}[!]
    \centering
    \includegraphics[trim={0 0 0 1.3cm},clip, width=0.6\textwidth]{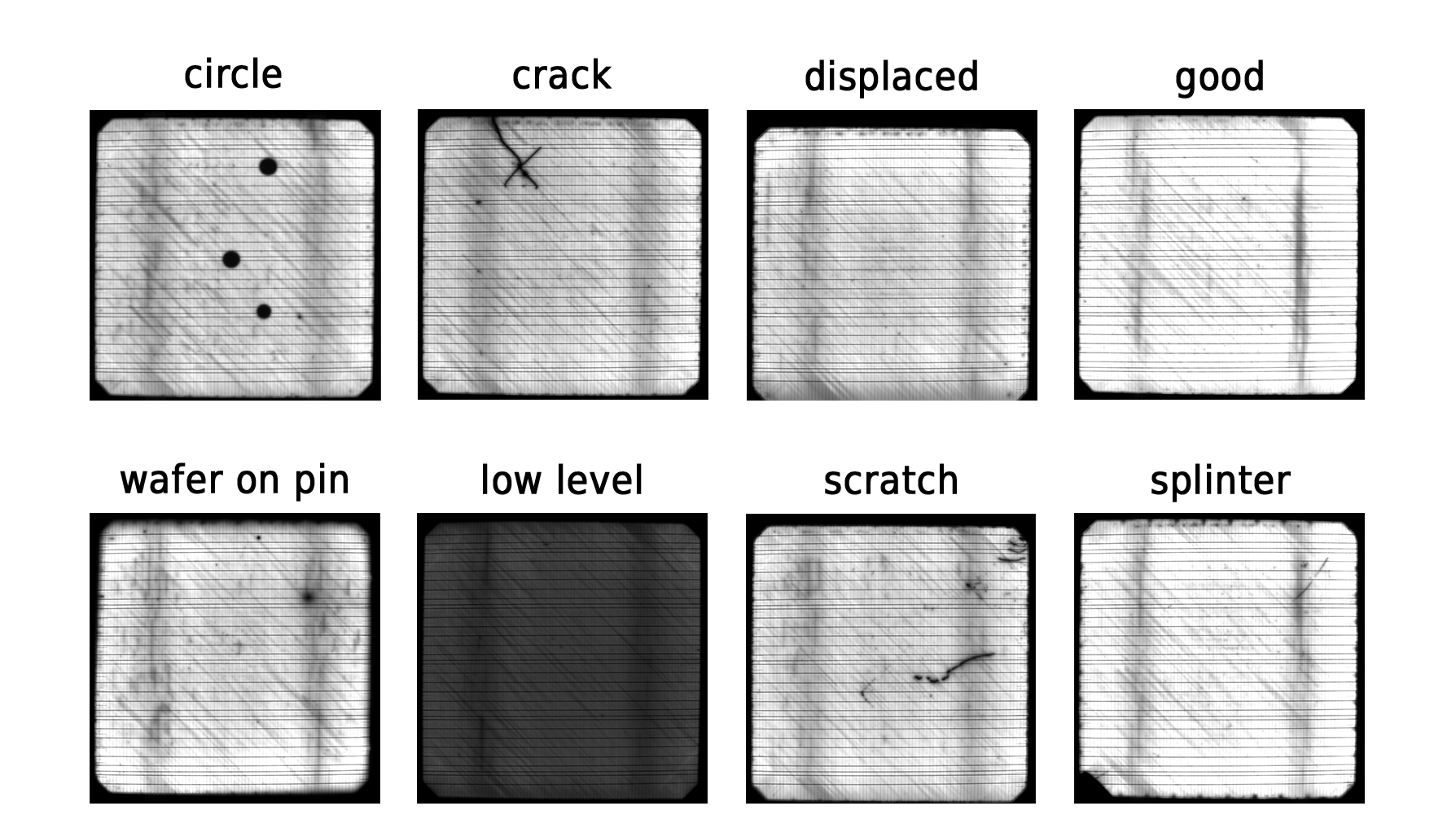}
    \caption{Example images for each of the eight classes.}
    \label{fig:dataset}
\end{figure}

In our case, a total of 4341 labeled images from a total of eight classes were captured by our partnering company. One class that denotes no errors, and a total of seven error-classes: low level, circle, crack, displaced, wafer on pin, splinter, and scratch. Apart from the self-explanatory classes, low level means that the glow of the wafers could not be recorded appropriately and wafer on pin means that only parts of the wafer are glowing. See Figure \ref{fig:dataset} for exemplary images of each class. 
We decided to generate three tasks from our dataset, comprising different sets of errors and, thus, varying complexity. We either used 3 classes (good, low level, circle - the classes that can be distinguished the easiest), 5 classes (good, low level, circle, crack, displaced), or all 8 classes (good, low level, circle, crack, displaced, waferonpin, splinter, scratch). The images were provided as TIFF-files with a resolution of 1024 x 1024 in 8bit grayscale. As Table \ref{tbl:descriptives} illustrates, the class distribution over the images is not balanced but differs quite heavily. 

\begin{table}[h]
\centering
\begin{tabular}{lcccc}
		    {\textbf{Class}} &
		    {\textbf{Number of}} & {\textbf{3 Classes}} & {\textbf{5 Classes}} & {\textbf{8 Classes}} \\
		    & {\textbf{observations}} \\
		\midrule
		Good & 1096 & \xmark & \xmark & \xmark \\
		Low level & 420 & \xmark & \xmark & \xmark \\
		Circle & 351 & \xmark & \xmark & \xmark \\
		Crack & 577 & & \xmark & \xmark \\
		Displaced & 993 & & \xmark & \xmark \\
		Wafer on pin & 256 & & & \xmark \\
		Splinter & 79 & & & \xmark \\
		Scratch & 569 & & & \xmark \\
		\midrule 
		$\sum$ & 4341 \\
\end{tabular}
\caption{Overview of our dataset and evaluation tasks.}
\label{tbl:descriptives}
\end{table}

\section{Experimental Setup}
\label{sec:experiments}

We conducted a series of computational experiments and compared the evaluation results systematically. Table \ref{tab:experiments} provides an overview of all experiments.

In line with our motivation to develop a lightweight yet high-performing model, we started with a simple \textit{baseline} approach in terms of a 2-layer CNN with two fully connected layers on top (ID 0). Building on this baseline, we continued with multiple experiments to increase the quality of our model, which can basically be divided into three different categories: 1) model-centric improvements, 2) data-centric improvements, and 3) a combination of both of them. 

The experiments on \textit{model-centric} improvements consider different architectural and hyperparameter-related design choices (ID 1-4). Further information is provided in Subsection \ref{sec:model-centric}. The experiments on \textit{data-centric} improvements mainly cover different data augmentation techniques (ID 5-7) and a color inversion approach (ID 8). Further details are provided in Subsection \ref{sec:data-centric}.
Since we wanted to exclude overlapping effects while also avoiding to check all combinations exhaustively due to high computational costs, the respective strands of model-centric and data-centric improvements were considered separately and \textit{combined} afterwards in case of promising effects (ID 9).

\begin{table}[h]
\begin{tabular}{lllll}
\textbf{}                                                                    & \textbf{Model} & \textbf{ID} & \textbf{Experiment}                                                                & \textbf{Description}                                                                                                                                                                                                          \\ \hline
Baseline                                                                     & BaseNet        & 0           & Baseline                                                                           & \begin{tabular}[c]{@{}l@{}}2 conv. layers ($5\times5$ kernel) \\ + ReLu + MaxPooling ($2\times2$ kernel) \\ + 2 FC layers\end{tabular}                                                                                       \\ \hline
\begin{tabular}[c]{@{}l@{}}Model-\\ centric\end{tabular}                     & BaseNet8       & 1           & \begin{tabular}[c]{@{}l@{}}Increase capacity \\ of BaseNet\end{tabular}            & \begin{tabular}[c]{@{}l@{}}8 conv. layers ($3\times3$ kernel) + ReLu\\ + MaxPooling ($2\times2$ kernel) \\ + 2 FC layers\end{tabular}                                                                                         \\
                                                                             & BaseNet8+      & 2           & Optimize learning                                                                  & \begin{tabular}[c]{@{}l@{}}BaseNet8 + ReLu6 + batch normalization \\ + multi-step learning\end{tabular}                                                                                                                       \\
                                                                             & IncNet         & 3           & \begin{tabular}[c]{@{}l@{}}Inception-like \\network\end{tabular} & \begin{tabular}[c]{@{}l@{}}4 serial custom Inception Blocks (paths: $1\times1$ \\ convolution (1),  and $3\times3$ (2), $5\times5$ (3), \\ and $9\times9$ (4) convolution followed by \\ $9\times9$ convolution)\end{tabular} \\
                                                                             & ResiNet        & 4           & \begin{tabular}[c]{@{}l@{}}ResNet-like \\ network\end{tabular}            & \begin{tabular}[c]{@{}l@{}}4 serial ResBlocks (paths: skip connection (1), \\ and two serial $3\times3$ convolutions (2))\end{tabular}                                                                                            \\ \hline
\begin{tabular}[c]{@{}l@{}}Data-\\ centric\end{tabular}                      & BaseNet        & 5           & \begin{tabular}[c]{@{}l@{}}Standard \\ Augmentations\end{tabular}                  & \begin{tabular}[c]{@{}l@{}}Horizontal and vertical flipping + 90° and 180° \\ rotations + Gaussian Blur + normalization\end{tabular}                                 \\
                                                                             &                & 6           & DeepSMOTE                                                                          & Generate feature-rich images using DeepSMOTE.                                                                                                                                                                              \\
                                                                             &                & 7           & Generated Images                                                                   & \begin{tabular}[c]{@{}l@{}}Randomly places prior masked errors onto good \\ wafer image; errors are positioned, resized,  \\ blurred, and rotated randomly.\end{tabular}                                                      \\
                                                                             &                & 8           & Color Invert                                                                      & Color inversion from white to black and vice versa.                                                                                                                                                                                                   \\ \hline
Comb.                                                                        & BaseNet8+      & 9           & \begin{tabular}[c]{@{}l@{}}BaseNet8+ \\ (+SA)\end{tabular}             & \begin{tabular}[c]{@{}l@{}}Combining the best results from model-centric \\  and data-centric experiments.\end{tabular}                                                                                                            \\ \hline
\multirow{3}{*}{\begin{tabular}[c]{@{}l@{}}Transfer\\ Learning\end{tabular}} & VGG16-FE       & 10          & \begin{tabular}[c]{@{}l@{}}VGG16 for FE\end{tabular}       & Fixing all weights except for classification head.                                                                                                                                                                             \\
                                                                             & VGG16-FT       & 11          & \begin{tabular}[c]{@{}l@{}}VGG16 with FT \end{tabular}         & \begin{tabular}[c]{@{}l@{}}Unfreezing all weights and fine-tuning them for\\our case.\end{tabular}                                                                                                                           \\
                                                                             & VGG16-FT       & 12          & \begin{tabular}[c]{@{}l@{}}VGG16 with FT\\ (+SA)\end{tabular}              & \begin{tabular}[c]{@{}l@{}}Combining VGG16-FT with our standard \\ augmentations.\end{tabular}                                                                                                                               
\end{tabular}
\caption{Overview of all experiments.}
\label{tab:experiments}
\end{table}

In order to assess all our "from-scratch-models" against the performance of \textit{transfer learning} models, we also conducted experiments with several versions of pre-trained models. More specifically, we used a VGG16 \cite{simonyan} that is pre-trained on the widely-known ImageNet dataset \cite{deng_imagenet_2009} and can classify images into 1,000 categories of generic objects. In a first version, the pre-trained VGG16 was used as a generic \textit{feature extractor} (FE; ID 10), and in a second version, the weights were adjusted via \textit{fine-tuning} (FT; ID 11) (cf. Subsection \ref{sec:CNN-TL}). Additionally, to ensure a fair comparison, we also added the best data-centric improvement to the best performing pre-trained model (ID 12).

\begin{figure}[!]
    \centering
    \includegraphics[width=.9\textwidth]{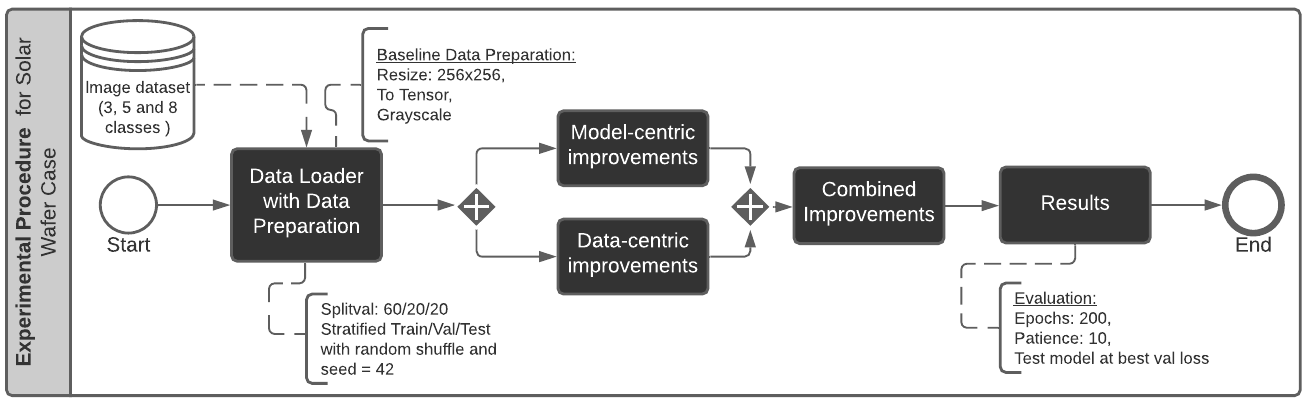}
    \caption{Model- and data-centric improvements procedure with common evaluation strategy.}
    \label{fig:pipeline}
\end{figure}

To provide better transparency about our overall experimental setup and the evaluation procedure, Figure \ref{fig:pipeline} summarizes our pipeline and the individual steps. As shown, we performed default data preparation steps before feeding the wafer images to the models in the corresponding experiments. These steps included resizing the images to a $256\times256$ resolution, casting the inputs to tensors, and converting the images to grayscales with one dimension. Additionally, the images for the pre-trained models were also normalized as recommended by the models. 

For our evaluation purposes, we use a stratified random split and divide our data into 60\% for training, 20\% for validation and early stopping, and 20\% for testing and the calculation of our final evaluation metrics: 1) precision, 2) recall, and 3) F1-score. Each metric is calculated with the weighted average across all classes as a common measure to assess the quality of multi-class classifiers \cite{heinrich_process_2021}. We use the F1-score as our main metric for comparisons. Additionally, we also measured the number of parameters of all trained models, the model size in megabyte (MB), and the inference times in milliseconds (ms). Across all experiments, we used the same data loading. Further, our training configurations are kept constant (cf. Appendix for details on implementation).


\subsection{Experiments on Model-centric Improvements}
\label{sec:model-centric}
As described above, we started with a simple, yet qualitative and performant model consisting of only two convolutional layers, called \textit{BaseNet} (ID 0). This model was iteratively extended to \textit{BaseNet8} (ID 1) through the addition of more layers and a larger number of filters. On top of that, we developed \textit{BaseNet8+} (ID 2), which is the same as BaseNet8 but with changed activation functions, batch normalization and additional hyperparameters. Beyond that, we also looked into CNN architectures with more advanced processing mechanisms (cf. \ref{sec:CNN-TL}). More specifically, we developed \textit{IncNet} (ID 3), inspired by Inception Networks \cite{szegedy_going_2015}, and \textit{ResiNet} (ID 4), inspired by ResNets \cite{he2016deep}. The configurations of the individual networks are summarized in Table \ref{tab:experiments}.

\subsection{Experiments on Data-centric Improvements}
\label{sec:data-centric}

In our data-centric experiments, the focus was on improving the data quality, while the underlying model remains the same. For this purpose, we used our BaseNet again for all experiments and evaluated it as described before.

We started with some \textit{standard augmentation} (SA) techniques (ID 5) to enhance the diversity of training data across all different classes, which has already proven successful in previous studies \cite{Gu2019ImproveIC}. These augmentations include horizontal and vertical flipping, as well as 90° and 180° rotations and Gaussian Blur. In addition, we also applied normalization by calculating mean and standard deviation of the training images.

Beyond that, two approaches were pursued to enhance the uneven distribution of the classes, i.e., data imbalance, by applying more specific data generation techniques. For this purpose, we chose two underrepresented classes with rather simple error structures, i.e., circles and splinters.
In a first approach, an \textit{automated image composition process} (ID 6) was followed to automatically generate \textit{n} images. In short, the algorithm selects a good wafer and randomly places prior masked errors onto the good wafer image. These  errors are then positioned, resized,  blurred and rotated based on randomness to achieve a high variability among the generated images.

\begin{figure}[h]
    \centering
    \includegraphics[width=0.6\textwidth]{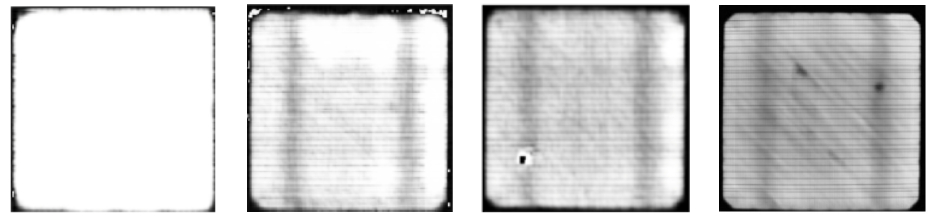}
    \caption{DeepSMOTE images over several iterations.}
    \label{fig:Random invert}
\end{figure}

In a second experiment, we considered a more innovative generation approach called \textit{Deep Synthetic Minority Over-sampling Technique} (DeepSMOTE) \cite{dablain_deepsmote_2021} (ID 7). Based on the traditional SMOTE technique \cite{chawla_smote_2002} embedded in a deep encoder and decoder architecture, DeepSMOTE generates feature-rich images to diminish disadvantages of unbalanced classes. Over several iterations, the encoder and decoder networks were optimized until the synthetically generated images showed the necessary information density. Figure \ref{fig:Random invert} shows some development stages up to the desired level of detail.

Finally, we investigated an approach that is rather domain-specific and might not be generalizable to other domains and datasets. We tested \textit{color inversion} such that the defects on the solar panel are now represented by bright/white pixels, which correspond to high numerical values. By relying on techniques such as MaxPooling those high values are amplified.








\section{Results}
\label{sec:results}
In this section, we present the results of our previously described experiments. We show that -- with a lightweight model -- we can achieve competitive performance compared to large pre-trained models consisting of 80$\times$ more parameters and other CNN architectures. Our main results are summarized in Table \ref{tab:5_intro}, in which we focus on the most complex case of recognizing images from eight different classes. Furthemore, we provide a complete overview of all results in Table \ref{tab:big_table} in our Appendix.

\begin{table}[!]
\begin{tabular}{l|llllllll}
\textbf{ID}               & \textbf{Model}                                                                & \textbf{Precision}           & \textbf{Recall}              & \textbf{F1-score}            & \textbf{Accuracy}            & \textbf{\#Params}                 & \textbf{Size (MB)}           & \textbf{\begin{tabular}[c]{@{}l@{}}Inference\\ Time (ms)\end{tabular}} \\ \hline
0                        & BaseNet                                                                       & 0.867                        & 0.871                        & 0.864                        & 0.871                        & 7,623,832                         & 35.74               & 0.44                                                          \\
\rowcolor[HTML]{FFFFFF} 
{\color[HTML]{000000} 3} & {\color[HTML]{000000} IncNet}                                                 & {\color[HTML]{000000} 0.943} & {\color[HTML]{000000} 0.944} & {\color[HTML]{000000} 0.943} & {\color[HTML]{000000} 0.944} & {\color[HTML]{000000} 61,740,488} & {\color[HTML]{000000} 360.5} & {\color[HTML]{000000} 10.97}                                           \\
{\color[HTML]{000000} 4} & {\color[HTML]{000000} ResiNet}                                                & {\color[HTML]{000000} 0.905} & {\color[HTML]{000000} 0.906} & {\color[HTML]{000000} 0.904} & {\color[HTML]{000000} 0.906} & {\color[HTML]{000000} 4,453,704}  & {\color[HTML]{000000} 48.93} & {\color[HTML]{000000} 1.74}                                            \\
\textbf{9}               & \textbf{\begin{tabular}[c]{@{}l@{}}BaseNet8+ (+SA)\\ \end{tabular}} & \textbf{0.971}               & \textbf{0.971}               & \textbf{0.971}               & \textbf{0.971}               & \textbf{1,707,208}      & \textbf{15.73}               & \textbf{0.50}                                                          \\
12                       & \begin{tabular}[c]{@{}l@{}}VGG16-FT (+SA) \\\end{tabular}           & 0.990                        & 0.990                        & 0.990                        & 0.990                        & 138,357,544                       & 814.18                       & 9,13                                                                  
\end{tabular}
\caption{Results of our lightweight BaseNet8+ (+SA) compared to competitor models.}
\label{tab:5_intro}
\end{table}

Starting from our simple BaseNet, it can be seen that the performance could be significantly elevated by our model-centric and data-centric improvements. The best performing model developed from scratch is our \textit{BaseNet8+ (+SA)} as a combination of a lightweight 8-layer CNN with improved configurations and standard data augmentations. With this combination, we could achieve a remarkable F1-score of 0.971. In contrast, the best pre-trained model achieved an F1-score of 0.99, which turned out to be \textit{VGG16-FT (+SA)} based on fine-tuning all weights and standard augmentations. Even though our best model is outperformed, it shows a huge parameter advantage by having 80$\times$ less parameters and is built without relying on training data from foreign image repositories. This also leads to the situation that our lightweight model requires far less storage capacities at 15.73 MB (52$\times$ smaller than VGG16-FT) and is extremely fast at 0.5 ms (18$\times$ faster than VGG16-FT). As alluded to earlier, these characteristics are crucial for applications in edge computing.

For further comparison, Table \ref{tab:5_intro} also contains the results of our ResiNet and IncNet models that were trained from scratch. Both approaches require a larger number of parameters, resulting in bigger model sizes and higher inference times, whereas the prediction qualities did not reach a satisfactory level. 




\subsection{Model-Centric Improvements, Transfer Learning, and Combined Models}
\label{sec:results_detail}

\begin{figure}[h]
    \centering
    \includegraphics[width=.9\textwidth]{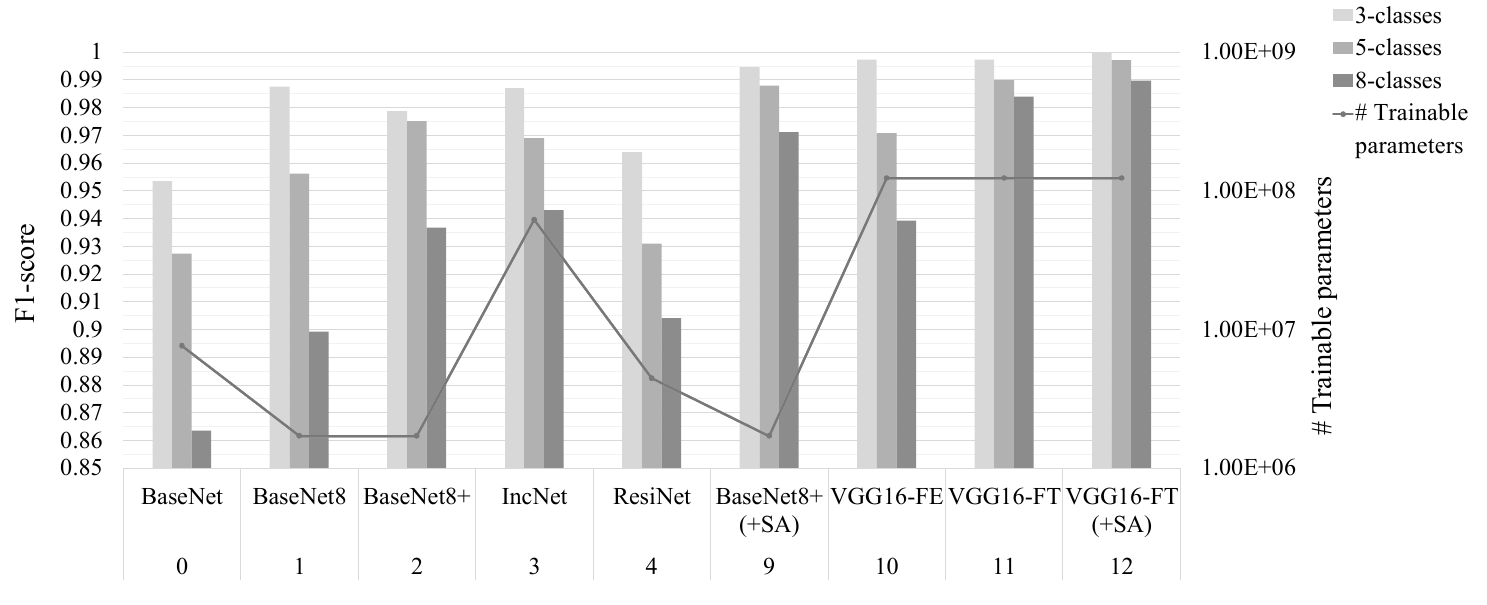}
    \caption{Results of model-centric, transfer learning and combined experiments.}
    \label{fig:model_centric_improvements}
\end{figure}

To provide further insights into the findings from our experiments, we summarize the results of the model-centric improvements, the transfer learning models, and the combined approaches in a joint diagram (cf. Figure \ref{fig:model_centric_improvements}). The data-centric experiments are considered separately in Subsection \ref{sec:results_data}. Our diagram shows the prediction qualities measured via F1-score using bar graphs for 3, 5, and 8 classes, respectively, whereas a line graph indicates the number of trainable model parameters on a logarithmic scale.

Focusing on the \text{model-centric experiments} (ID 1-4), we can see a steady improvement from the BaseNet to BaseNet8 to BaseNet8+ to IncNet. ResiNet performs worse than BaseNet8+ and IncNet, while still having $2.6\times$ more parameters than BaseNet8+. ResiNet and IncNet have a significantly higher number of parameters ($36\times$ and $2.6\times$) than our BaseNet models. Due to this huge difference in number of parameters, and ResiNet performing similar to BaseNet8 and worse than BaseNet8+, as well as IncNet having the slowest inference time of all models, we decided not to pursue these two architectures any further. Instead, we added the best performing data-centric techniques in terms of standard augmentations (cf. Subsection \ref{sec:results_data}) (ID 9), which could further boost the prediction qualities by a considerable margin across all three complexity levels for our best performing model BaseNet8+ (+SA).

When focusing on the results from \textit{transfer learning}, we can see that all pre-trained VGG16 models (ID 10-12) generally achieve the best prediction qualities across the different complexity levels. It is notable that the VGG16-FE as generic feature extractor struggles to detect the domain-specific shapes of the different defect types on the solar panels, which is reflected by the inferior performance for five classes (0.971) and eight classes (0.939), respectively. However, once all the weights of the pre-trained model are adjusted via fine-tuning on the solar images within the VGG16-FT experiment, the model achieves outstanding results above 0.98 for all three complexity levels. Through additional standard augmentation, these results could even be raised to a performance level of over 0.99 for all classes.

Nevertheless, considering the drawbacks of these large models in terms of number of parameters, huge model sizes and high inference times, it is striking that our lightweight BaseNet8+ (+SA) model with far fewer layers and parameters comes very close to these results. As such, it can even outperform the results of VGG16-FE on five classes (0.988 vs. 0.971) and eight classes (0.971 vs. 0.939), respectively, which is also the case for our BaseNet8+ without augmentation on five classes (0.975 vs. 0.971).



\subsection{Data-centric Improvements}
\label{sec:results_data}

\begin{figure}[h]
    \centering
    \includegraphics[width=.9\textwidth]{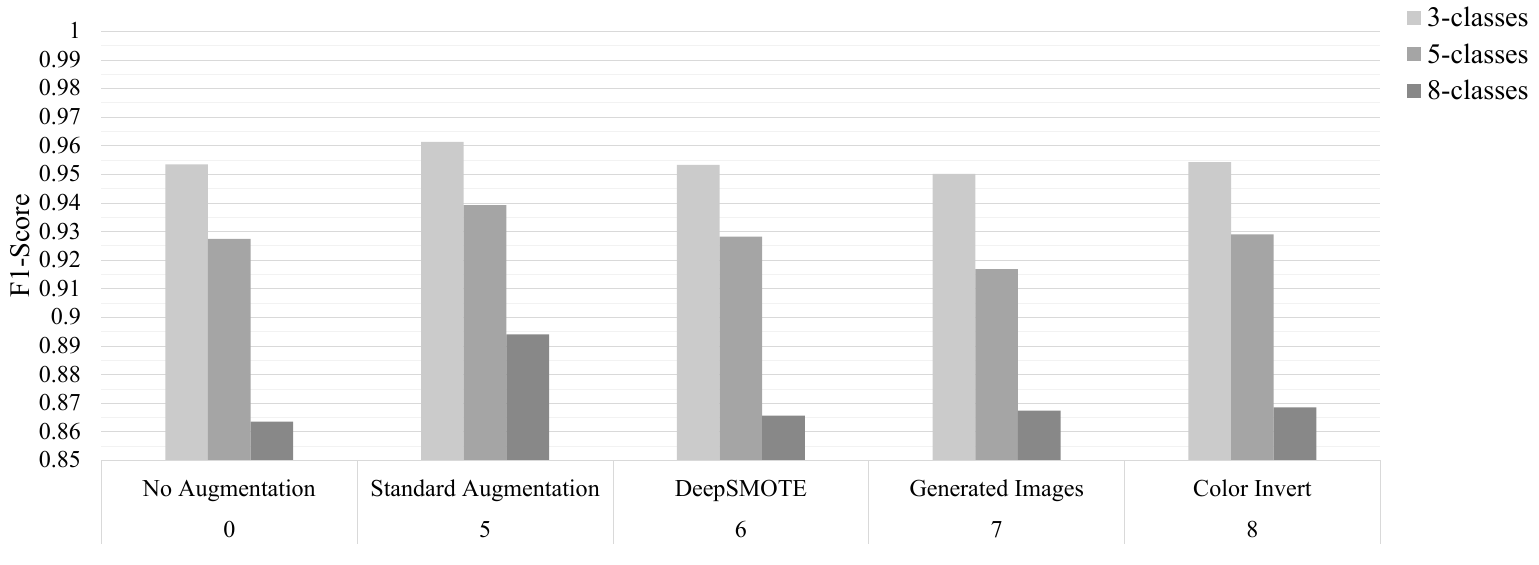}
    \caption{Results of data-centric experiments with fixed BaseNet model.}
    \label{fig:data_centric}
\end{figure}

Considering the results of the data-centric experiments, the improvements across all implemented approaches are rather limited. Figure \ref{fig:data_centric} summarizes the results.

It can be seen that -- compared to the baseline (ID 0) -- \textit{standard augmentations} (ID 5) led to the greatest overall improvement of approximately 0.017 (F1-score), as the transformations increase the stability of the model. Comparing the two more innovative/ sophisticated augmentation methods, both did not lead to a significant improvement. On average over the three class groupings, \textit{DeepSMOTE} (ID 6) increased the performance only by 0.001. However, it reflects characteristics of the existing data better than the \textit{generated images} which even worsened the performance by 0.003 - most probably because they introduce patterns that do not correspond to the real data. Finally, the domain-specific approach of using color inversion led to an improvement of only 0.002 on average. As all data-centric approaches, besides standard augmentation, did not show promising improvements, only standard augmentation was used for our combined experiments.

\section{Discussion and Concluding Remarks}
\label{sec:conclusion}

Industrial computer vision greatly benefits from recent developments in the field of deep neural networks. In particular, transfer learning and pre-training have allowed the usage of huge neural networks to be applied to all kinds of vision-based tasks. However, these large pre-trained DNNs are not applicable in many cases, such as in edge computing where small devices detect errors on products on the shop floor level of companies. 

To this end, we studied small, hand-crafted CNNs which are specifically powerful in their fast inference times and high accuracies. Thereby, we contribute to the field of CV, but also to the thriving field of analytics in edge computing. Exemplary, we used a case from the PV industry, in which cost-effective optimization of the value chains promises large benefits. However, our approaches are not limited to this field but can be applied to a manifold of challenges, where fast inference times on small devices are crucial.

This work is not free of limitations. In contrast to randomly exploring the search space of CNN architectures (e.g., through automated hyperparameter optimization), we decided to strategically analyse in two directions: model-centric and data-centric approaches. For the latter, we decided to try four different variations, yet, want to highlight that a manifold of other data-centric approaches are appealing. For instance, to improve the DeepSMOTE results, a further increase of the capacity of the encoder and decoder architecture could potentially improve the degree of details of generated images.  Alternatively, generative adversarial networks (GANs) \cite{goodfellow_generative_2014} might be a valuable approach to achieve a higher degree of details. Similarly, we did not evaluate all feasible combinations of our data-centric and model-centric approaches which delimits the generalizability of our findings to some extent. Furthermore, we need to diagnose our models to verify their robustness, for which we intend to use several XAI techniques \cite{barredo_arrieta_explainable_2020} in subsequent studies.

We expect that the following years will demand for more research on edge analytics. This means that more research will be needed on small, yet powerful ANNs for industry cases. This work is a first step in that direction. We exploit recent ideas in the field of CV to improve performance of specifically small models and show that our models are tough competitors for large DNNs with several hundreds of millions of parameters. In the future, we want to verify the robustness of our models in comparison to pre-trained models using XAI techniques, extend our experiments using DeepSMOTE, and want to test our approach for generalization over other datasets.

\section*{Appendix}
\label{sec:appendix}
\addcontentsline{toc}{section}{Appendix}
\renewcommand{\thesubsection}{\Alph{subsection}}

\label{app:baseline_hyperparameter}
Our configurations/hyperparameters were kept constant throughout all experiments, which were performed on Google Colab (standard version): Optimizer = Adam optimizer; number of epochs = 200; batch size = 64; weight decay = 0.0001; early stopping patience = 10; learning rate = slightly varying between 0.001 and 0.0001. Our results and implementations can be found here: \url{https://github.com/maximus-victor/DViP4WI22}






\label{sec:detailed_res}
\begin{table}[!]
\begin{tabular}{lll|llllll|lll}
\textbf{Approach}                                                                     & \textbf{Model}                      & \textbf{Config}                                                                                                                                                  & \textbf{ID}         & \textbf{Cl.}       & \textbf{Prec.}         & \textbf{Recall}        & \textbf{F1}            & \textbf{Acc.}          & \textbf{\#Params}                     & \textbf{\begin{tabular}[c]{@{}l@{}}Size\\ (MB)\end{tabular}} & \textbf{\begin{tabular}[c]{@{}l@{}}Inf.\\ Time\\ (ms)\end{tabular}} \\ \hline
\multirow{3}{*}{\textbf{Baseline}}                                                    & \multirow{3}{*}{\textbf{BaseNet}}   & \multirow{3}{*}{\textbf{\begin{tabular}[c]{@{}l@{}}2-Layer, \\ Kernel-Size 5,\\ 8 Kernels\end{tabular}}}                                                         & \textbf{0a}         & \textbf{3}         & \textbf{0.955}         & \textbf{0.955}         & \textbf{0.953}         & \textbf{0.955}         & \multirow{3}{*}{\textbf{7,623,832}}   & \multirow{3}{*}{\textbf{35.74}}                              & \multirow{3}{*}{\textbf{0.44}}                                      \\
                                                                                      &                                     &                                                                                                                                                                  & \textbf{0b}         & \textbf{5}         & \textbf{0.928}         & \textbf{0.931}         & \textbf{0.927}         & \textbf{0.931}         &                                       &                                                              &                                                                     \\
                                                                                      &                                     &                                                                                                                                                                  & \textbf{0c}         & \textbf{8}         & \textbf{0.867}         & \textbf{0.871}         & \textbf{0.864}         & \textbf{0.871}         &                                       &                                                              &                                                                     \\ \hline
\multirow{15}{*}{\begin{tabular}[c]{@{}l@{}}Model-\\ centric\end{tabular}}            & \multirow{3}{*}{BaseNet8}           & \multirow{3}{*}{\begin{tabular}[c]{@{}l@{}}8-Layer, \\ Kernel-Size 3, \\ 512 Kernels\end{tabular}}                                                               & 1a                  & 3                  & 0.988                  & 0.988                  & 0.988                  & 0.988                  & \multirow{3}{*}{1,707,208}            & \multirow{3}{*}{15.73}                                       & \multirow{3}{*}{0.50}                                               \\
                                                                                      &                                     &                                                                                                                                                                  & 1b                  & 5                  & 0.957                  & 0.957                  & 0.956                  & 0.957                  &                                       &                                                              &                                                                     \\
                                                                                      &                                     &                                                                                                                                                                  & 1c                  & 8                  & 0.899                  & 0.902                  & 0.899                  & 0.902                  &                                       &                                                              &                                                                     \\ \cline{2-12} 
                                                                                      & \multirow{6}{*}{BaseNet8+}          & \multirow{6}{*}{\begin{tabular}[c]{@{}l@{}}8-Layer, \\ Kernel-Size 3, \\ 512 Kernels, \& \\ Activation Layer, \\ MultiStep Scheduler, \\ BatchNorm\end{tabular}} & \multirow{2}{*}{2a} & \multirow{2}{*}{3} & \multirow{2}{*}{0.979} & \multirow{2}{*}{0.979} & \multirow{2}{*}{0.979} & \multirow{2}{*}{0.979} & \multirow{6}{*}{1,707,208}            & \multirow{6}{*}{15.73}                                       & \multirow{6}{*}{0.50}                                               \\
                                                                                      &                                     &                                                                                                                                                                  &                     &                    &                        &                        &                        &                        &                                       &                                                              &                                                                     \\
                                                                                      &                                     &                                                                                                                                                                  & \multirow{2}{*}{2b} & \multirow{2}{*}{5} & \multirow{2}{*}{0.975} & \multirow{2}{*}{0.975} & \multirow{2}{*}{0.975} & \multirow{2}{*}{0.975} &                                       &                                                              &                                                                     \\
                                                                                      &                                     &                                                                                                                                                                  &                     &                    &                        &                        &                        &                        &                                       &                                                              &                                                                     \\
                                                                                      &                                     &                                                                                                                                                                  & \multirow{2}{*}{2c} & \multirow{2}{*}{8} & \multirow{2}{*}{0.937} & \multirow{2}{*}{0.937} & \multirow{2}{*}{0.937} & \multirow{2}{*}{0.937} &                                       &                                                              &                                                                     \\
                                                                                      &                                     &                                                                                                                                                                  &                     &                    &                        &                        &                        &                        &                                       &                                                              &                                                                     \\ \cline{2-12} 
                                                                                      & \multirow{3}{*}{IncNet}             & \multirow{3}{*}{basic configuration}                                                                                                                             & 3a                  & 3                  & 0.987                  & 0.987                  & 0.987                  & 0.987                  & \multirow{3}{*}{61,740,488}           & \multirow{3}{*}{360.5}                                       & \multirow{3}{*}{10.97}                                              \\
                                                                                      &                                     &                                                                                                                                                                  & 3b                  & 5                  & 0.969                  & 0.969                  & 0.969                  & 0.969                  &                                       &                                                              &                                                                     \\
                                                                                      &                                     &                                                                                                                                                                  & 3c                  & 8                  & 0.943                  & 0.944                  & 0.943                  & 0.944                  &                                       &                                                              &                                                                     \\ \cline{2-12} 
                                                                                      & \multirow{3}{*}{ResiNet}            & \multirow{3}{*}{basic configuration}                                                                                                                             & 4a                  & 3                  & 0.966                  & 0.965                  & 0.964                  & 0.965                  & \multirow{3}{*}{4,453,704}            & \multirow{3}{*}{48.93}                                       & \multirow{3}{*}{1.74}                                               \\
                                                                                      &                                     &                                                                                                                                                                  & 4b                  & 5                  & 0.932                  & 0.935                  & 0.931                  & 0.935                  &                                       &                                                              &                                                                     \\
                                                                                      &                                     &                                                                                                                                                                  & 4c                  & 8                  & 0.905                  & 0.906                  & 0.904                  & 0.906                  &                                       &                                                              &                                                                     \\ \hline
\multirow{12}{*}{\begin{tabular}[c]{@{}l@{}}Data-\\ centric\end{tabular}}             & \multirow{12}{*}{BaseNet}           & \multirow{3}{*}{Standard Aug.}                                                                                                                                   & 9a                  & 3                  & 0.962                  & 0.962                  & 0.961                  & 0.962                  & \multicolumn{3}{c}{\multirow{12}{*}{see BaseNet}}                                                                                                                          \\
                                                                                      &                                     &                                                                                                                                                                  & 9b                  & 5                  & 0.942                  & 0.943                  & 0.939                  & 0.943                  & \multicolumn{3}{c}{}                                                                                                                                                       \\
                                                                                      &                                     &                                                                                                                                                                  & 9c                  & 8                  & 0.896                  & 0.897                  & 0.894                  & 0.897                  & \multicolumn{3}{c}{}                                                                                                                                                       \\ \cline{3-9}
                                                                                      &                                     & \multirow{3}{*}{DeepSMOTE}                                                                                                                                       & 6a                  & 3                  & 0.953                  & 0.954                  & 0.953                  & 0.954                  & \multicolumn{3}{c}{}                                                                                                                                                       \\
                                                                                      &                                     &                                                                                                                                                                  & 6b                  & 5                  & 0.928                  & 0.929                  & 0.928                  & 0.929                  & \multicolumn{3}{c}{}                                                                                                                                                       \\
                                                                                      &                                     &                                                                                                                                                                  & 6c                  & 8                  & 0.867                  & 0.871                  & 0.866                  & 0.871                  & \multicolumn{3}{c}{}                                                                                                                                                       \\ \cline{3-9}
                                                                                      &                                     & \multirow{3}{*}{Generated Images}                                                                                                                                & 7a                  & 3                  & 0.952                  & 0.952                  & 0.950                  & 0.952                  & \multicolumn{3}{c}{}                                                                                                                                                       \\
                                                                                      &                                     &                                                                                                                                                                  & 7b                  & 5                  & 0.919                  & 0.922                  & 0.917                  & 0.922                  & \multicolumn{3}{c}{}                                                                                                                                                       \\
                                                                                      &                                     &                                                                                                                                                                  & 7c                  & 8                  & 0.868                  & 0.873                  & 0.867                  & 0.873                  & \multicolumn{3}{c}{}                                                                                                                                                       \\ \cline{3-9}
                                                                                      &                                     & \multirow{3}{*}{Color Invert}                                                                                                                                    & 8a                  & 3                  & 0.955                  & 0.955                  & 0.954                  & 0.955                  & \multicolumn{3}{c}{}                                                                                                                                                       \\
                                                                                      &                                     &                                                                                                                                                                  & 8b                  & 5                  & 0.929                  & 0.932                  & 0.929                  & 0.932                  & \multicolumn{3}{c}{}                                                                                                                                                       \\
                                                                                      &                                     &                                                                                                                                                                  & 8c                  & 8                  & 0.871                  & 0.873                  & 0.869                  & 0.873                  & \multicolumn{3}{c}{}                                                                                                                                                       \\ \hline
\multirow{3}{*}{\textbf{Comb.}}                                                       & \multirow{3}{*}{\textbf{BaseNet8+}} & \multirow{3}{*}{\textbf{\begin{tabular}[c]{@{}l@{}}BaseNet8+ \\ (+SA)\end{tabular}}}                                                                             & \textbf{9a}         & \textbf{3}         & \textbf{0.995}         & \textbf{0.995}         & \textbf{0.995}         & \textbf{0.995}         & \multirow{3}{*}{\textbf{1,707,208}}   & \multirow{3}{*}{\textbf{15.73}}                              & \multirow{3}{*}{\textbf{0.50}}                                      \\
                                                                                      &                                     &                                                                                                                                                                  & \textbf{9b}         & \textbf{5}         & \textbf{0.988}         & \textbf{0.988}         & \textbf{0.988}         & \textbf{0.988}         &                                       &                                                              &                                                                     \\
                                                                                      &                                     &                                                                                                                                                                  & \textbf{9c}         & \textbf{8}         & \textbf{0.971}         & \textbf{0.971}         & \textbf{0.971}         & \textbf{0.971}         &                                       &                                                              &                                                                     \\ \hline
\multirow{9}{*}{\textbf{\begin{tabular}[c]{@{}l@{}}Transfer\\ Learning\end{tabular}}} & \multirow{3}{*}{VGG16-FE}           & \multirow{3}{*}{\begin{tabular}[c]{@{}l@{}}VGG16 only for \\ feature extraction\end{tabular}}                                                                    & 10a                 & 3                  & 0.997                  & 0.997                  & 0.997                  & 0.997                  & \multirow{9}{*}{\textbf{138,357,544}} & \multirow{9}{*}{\textbf{814.18}}                             & \multirow{9}{*}{\textbf{9.13}}                                      \\
                                                                                      &                                     &                                                                                                                                                                  & 10b                 & 5                  & 0.972                  & 0.971                  & 0.971                  & 0.971                  &                                       &                                                              &                                                                     \\
                                                                                      &                                     &                                                                                                                                                                  & 10c                 & 8                  & 0.941                  & 0.940                  & 0.939                  & 0.940                  &                                       &                                                              &                                                                     \\ \cline{2-9}
                                                                                      & \multirow{6}{*}{\textbf{VGG16-FT}}  & \multirow{3}{*}{\begin{tabular}[c]{@{}l@{}}VGG16 with all\\ weights fine-tuned\end{tabular}}                                                                      & 11a                 & 3                  & 0.997                  & 0.997                  & 0.997                  & 0.997                  &                                       &                                                              &                                                                     \\
                                                                                      &                                     &                                                                                                                                                                  & 11b                 & 5                  & 0.990                  & 0.990                  & 0.990                  & 0.990                  &                                       &                                                              &                                                                     \\
                                                                                      &                                     &                                                                                                                                                                  & 11c                 & 8                  & 0.984                  & 0.984                  & 0.984                  & 0.984                  &                                       &                                                              &                                                                     \\ \cline{3-9}
                                                                                      &                                     & \multirow{3}{*}{\textbf{\begin{tabular}[c]{@{}l@{}}VGG16-FT \\ (+SA)\end{tabular}}}                                                                              & \textbf{12a}        & \textbf{3}         & \textbf{1}             & \textbf{1}             & \textbf{1}             & \textbf{1}             &                                       &                                                              &                                                                     \\
                                                                                      &                                     &                                                                                                                                                                  & \textbf{12b}        & \textbf{5}         & \textbf{0.997}         & \textbf{0.997}         & \textbf{0.997}         & \textbf{0.997}         &                                       &                                                              &                                                                     \\
                                                                                      &                                     &                                                                                                                                                                  & \textbf{12c}        & \textbf{8}         & \textbf{0.990}         & \textbf{0.990}         & \textbf{0.990}         & \textbf{0.990}         &                                       &                                                              &                                                                    
\end{tabular}
\caption{Overall results of all experiments (baseline and best models in bold).}
\label{tab:big_table}
\end{table}




\bibliographystyle{splncs04}
\bibliography{literature.bib}



\end{document}